\documentclass[10pt]{article}   % LaTeX2e

\usepackage[preprint]{rlj}        % de-anonymized preprint
\usepackage[dvipsnames]{xcolor}
\usepackage{amssymb}            % defines common symbols like \mathbb R
\usepackage{algorithm}          %
\usepackage{algpseudocode}      % wrapper over algorithmicx
\usepackage{bm}                 % bold math
\usepackage{graphicx}           % including images
\usepackage{mathtools}          % extends amsmath, providing common math tools
\usepackage{mathrsfs}           % enables \mathscr, when \mathcal does not work
\usepackage{subcaption}         % allows for the use of subfigures and sub-captions
\usepackage{url}                % displaying URLs
\usepackage[normalem]{ulem}
\usepackage{booktabs}
\usepackage{acronym}            % defines acronyms
\usepackage[framemethod=TikZ]{mdframed}  % display boxes
\usepackage{wrapfig}            % wrap figures
\usepackage{siunitx}            % nicely formatted scientific notation and units
\usepackage[space]{grffile}     % spaces in image names

\usepackage{lipsum}             % placeholder text
\usepackage{todonotes}          % margin comments

\usepackage{math0}
\usepackage{math1}

\graphicspath{{./img}}          % set the image path
\mathtoolsset{showonlyrefs}     % only number equations that are referenced

\definecolor{red}{HTML}{f8766d}
\definecolor{green}{HTML}{7cae00}
\definecolor{blue}{HTML}{00bfc4}
\definecolor{purple}{HTML}{c77cff}
\definecolor{black}{HTML}{14141c}
\definecolor{mgray}{HTML}{f1f4f7}

\mdfdefinestyle{leadingQ}{%
  linecolor=white,%
  linewidth=0,%
  backgroundcolor=mgray,%
  skipabove=\topsep,%
  skipbelow=0,%
  innerleftmargin=10pt,%
  innerrightmargin=10pt,%
  innerbottommargin=0.5\baselineskip,%
  innertopmargin=0.5\baselineskip,%
  roundcorner=4pt,%
  font=\normalsize\bfseries,%
  nobreak=true,%
}
\NewDocumentCommand{\strikeAndWrite}{ m o m }{%
  \IfValueTF{#2}{\sout{#2}}{} #1{#3}%
}
\NewDocumentCommand{\razp}{o m}{\strikeAndWrite{\textcolor{red}}[#1]{#2}}
\NewDocumentCommand{\adi}{o m}{\strikeAndWrite{\textcolor{green}}[#1]{#2}}
\NewDocumentCommand{\florin}{o m}{\strikeAndWrite{\textcolor{blue}}[#1]{#2}}

\NewDocumentCommand{\defineMarginNote}{mm}{
  \expandafter\NewDocumentCommand\csname#1\endcsname {mO{}}
  {%
    \todo[linecolor=#2,bordercolor=#2,backgroundcolor=#2!20,##2]{##1}%
  }
}
\defineMarginNote{Razp}{red}
\defineMarginNote{Adi}{green}
\defineMarginNote{Florin}{blue}
\newcommand{\adam}{\textsc{Adam}}
\newcommand{\aql}{\textsc{AQ}$(\lambda)$}
\newcommand{\atari}{\textsc{Atari}}
\newcommand{\atariFive}{\textsc{Atari-5$^\ast$}}
\newcommand{\dqn}{\textsc{DQN}}
\newcommand{\cdqn}{\textsc{C51}}
\newcommand{\minatar}{\textsc{MinAtar}}
\newcommand{\rmsprop}{RMSprop}
\newcommand{\streamq}{\textsc{StreamQ}}
\newcommand{\tdzero}{TD(0)}
\newcommand{\numGames}{55}

\algnewcommand{\LineComment}[1]{\Statex {\(\triangleright\) #1}}  % line comment in algo

\acrodef{adam}[\textsc{Adam}]{Adaptive Moment Estimation}
\acrodef{ale}[ALE]{Arcade Learning Environment}
\acrodef{aql}[\textsc{AQ}$(\lambda)$]{Adaptive $Q(\lambda)$}
\acrodef{c51}[\textsc{C51}]{Categorical DQN}
\acrodef{dqn}[\textsc{DQN}]{Deep Q-Network}
\acrodef{drl}[DRL]{deep reinforcement learning}
\acrodef{mdp}[MDP]{Markov decision process}
\acrodef{qr}[QR]{Quantile Regression}
\acrodef{qrc}[QRC]{Q-Learning with Regularized Corrections}
\acrodef{rl}[RL]{reinforcement learning}
\acrodef{rmsprop}[RMSProp]{Root Mean Square Propagation}
\acrodef{sgdm}[\textsc{SGD-M}]{stochastic gradient descent with momentum}
\acrodef{sq}[\textsc{StreamQ}]{StreamQ}

\title{Revisiting Adam for Streaming Reinforcement\\ Learning}
\setrunningtitle{Revisiting Adam for Streaming Reinforcement Learning}

\author{%
  Florin Gogianu\textsuperscript{1},
  Adrian Catalin Lutu\textsuperscript{1},
  Razvan Pascanu\textsuperscript{2}
}

\emails{florin.gogianu@gmail.com}

\affiliations{
$^1$\textbf{Bitdefender, Bucharest, Romania},
$^2$\textbf{Mila, Montreal, Canada}
}

\contribution{
  We highlight the effectiveness of \adam\ and established objective functions in the streaming RL setting.
}{
  Recent literature suggests that the severe non-stationarity of streaming environments requires specialized algorithms. We contrast this by adapting \dqn\ and \cdqn\ to the streaming setup, showing that when appropriately tuned, standard optimisation techniques can be surprisingly effective.
}

\contribution{
  We establish variance-adjusted optimisation algorithms, together with carefully tuned $\varepsilon$-values, to be important components in streaming RL agents.
}{
  We provide mechanistic insights into the role of variance-adjusted updates and \adam's epsilon value.
  We bring evidence that large $\varepsilon$ values can act as a filter for sparse and noisy gradient components, allowing for a more stable learning dynamics.
}

\contribution{
  We propose Adaptive $Q(\lambda)$.
}{
  $AQ(\lambda)$ combines eligibility traces with the variance adaptation mechanism of \adam\ and bounded error signal, resulting in an algorithm which improves upon the performance of existing streaming RL methods.
}

\contribution{
  We perform a large empirical evaluation in the streaming RL setting, extensively benchmarking \streamq\ alongside \dqn\ and \cdqn\ across \numGames\ Atari games and setting new performance expectations for these algorithms.
}{
  To provide a rigorous demonstration of our algorithmic contributions, the \numGames\ Atari games were selected and ordered based on the performance correlation framework established by \cite{aitchison2023atari5}.
}


\summary{
  Learning from a sequence of interactions, as soon as observations are perceived and acted upon, without explicitly storing them, holds the promise of simpler, more efficient and adaptive algorithms.
  For over a decade, however, deep reinforcement learning walked the contrary path, augmenting agents with replay buffers or parallel sampling routines, in an effort to tame learning instability.
  Recently, this topic has been revisited by \cite{elsayed2024streamin}, focusing on update computation through eligibility traces and modifications to the optimisation routine, resulting in the StreamQ algorithm.
  In this work we take a step back, investigating the efficacy of established updates, such as those implemented by DQN and C51 within this online setting.
  Not only do we find that they perform well, but through analysing how the optimisation algorithm generally, and Adam in particular, interacts with these updates, we contend that two properties are essential for robust performance:
  i) the derivative of the objective is to be bounded and ii) weight updates are variance-adjusted.
  Rigorous and exhaustive experimentation demonstrates that C51, which exhibits both characteristics, is competitive with StreamQ across a subset of \numGames\ Atari games.
  Using these insights, we derive a variance-adjusted algorithm based on eligibility traces, termed Adaptive Q$(\lambda)$, which approaches double the human baseline on the same subset, surpassing existing methods by all performance metrics.
}

\begin{document}

\makeCover    % make the cover page
\maketitle      % make the title section

\begin{abstract}
  Learning from a sequence of interactions, as soon as observations are perceived and acted upon, without explicitly storing them, holds the promise of simpler, more efficient and adaptive algorithms.
  For over a decade, however, deep reinforcement learning walked the contrary path, augmenting agents with replay buffers or parallel sampling routines, in an effort to tame learning instability.
  Recently, this topic has been revisited by \cite{elsayed2024streamin}, focusing on update computation through eligibility traces and modifications to the optimisation routine, resulting in the StreamQ algorithm.
  In this work we take a step back, investigating the efficacy of established updates, such as those implemented by DQN and C51 within this online setting.
  Not only do we find that they perform well, but through analysing how the optimisation algorithm generally, and Adam in particular, interacts with these updates, we contend that two properties are essential for robust performance:
  i) the derivative of the objective is to be bounded and ii) weight updates are variance-adjusted.
  Rigorous and exhaustive experimentation demonstrates that C51, which exhibits both characteristics, is competitive with StreamQ across a subset of 55 Atari games.
  Using these insights, we derive a variance-adjusted algorithm based on eligibility traces, termed Adaptive Q$(\lambda)$, which approaches double the human baseline on the same subset, surpassing existing methods by all performance metrics.
\end{abstract}

\section{Introduction}
\label{sec:intro}

Much of the performance gains in \ac{drl} over the past decade can be traced back to scaling the amount of off-policy data the optimisation routine can use during training.
Ever-increasing experience replay buffers or massively parallel sampling of the environment, allow stochastic gradient descent algorithms originally developed for empirical risk minimisation to play for their strengths, at the cost of large memory, sample and algorithmic complexity.
It became rather the rule that we tackled environments that fitted in 4Kb of \textsc{ROM} with agents consisting of 7GB replay buffers and millions of parameters in an effort to meet the stochastic gradient descent requirements for stability \citep{bowling2025talk}.

\begin{figure}[htbp]
  \centering
  \begin{subfigure}[t]{0.666\linewidth}
    \centering
    \includegraphics[width=\linewidth]{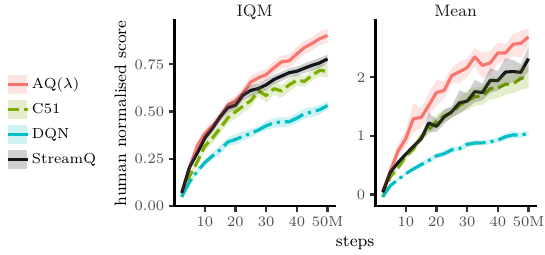}
  \end{subfigure}%
  \hfill
  \begin{subfigure}[t]{0.333\linewidth}
    \centering
    \includegraphics[width=\linewidth]{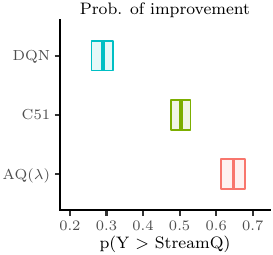}
  \end{subfigure}
  \caption{IQM and Mean human normalised score on  \numGames\ \atari\ games.
  All agents learn solely from the current transition, without experience replay, or target network.
  \acs{c51} and \acs{aql} demonstrate the usefulness of bounded objectives and variance-adjusted optimisation.
  }
  \label{fig:main}
\end{figure}

% % \begin{wrapfigure}{r}{0.5\textwidth}
% \begin{figure}
%   % \vspace{-\baselineskip}
%   \begin{center}
%     \includegraphics{ale_tc_iqm_mean_acc.pdf}
%     % \includegraphics{ale_tc_iqm_mean.pdf}
%   \end{center}
%   % \vspace{-10pt}
%   \caption{IQM and Mean human normalised score on  \numGames\ \atari\ games.
%   %
%   All agents learn solely from the current transition, without experience replay, or target network.
%   %
%   \acs{c51} and \acs{aql} demonstrate the usefulness of bounded objectives and variance-adjusted optimisation.
%   %
%   \florin{Wrap it or add facets and full row}
%   }
%   % \vspace{-16pt}
%   \label{fig:ale-iqm-mean}
% \end{figure}
% % \end{wrapfigure}

In contrast, few voices in the community started to argue for developing computationally bounded agents, with a reduced complexity compared to that of the environment \citep{javed2024bigWorld,bowling2025rethinking,lewandowski2025bigWorld}.
Thus, moving away from learning with batches sampled from pseudo-stationary buffers and towards learning from each interaction at a time, in a streaming fashion, without storing data or leveraging parallel environments.

Arguably, a major desideratum for this vision is the development of new RL objectives and optimisation algorithms that learn stably and efficiently purely from sequential data.
Inroads have been opened in by \cite{seijen2016trueOnlineTD,vanHasselt2014trueGTDlmbda,javed2024swiftTD} and recent results by \cite{elsayed2024streamin,elelimy2025gradient} suggest a way forward even for historically difficult control benchmarks such as the \ac{ale}.
These recent breakthroughs in the online setting have been attributed to the resurgence of eligibility traces \citep{sutton1988tdlmbda,sutton2018rlbook} based methods \citep{vanHasselt2014trueGTDlmbda,mahmood2015wisTrace,white2016greedy,vanHasselt2021expected}.

Indeed, eligibility traces for long held the promise of solving the central problem of propagating present or future rewards to past states and actions by updating value functions at every transition using multistep returns.
Improved temporal credit assignment would in turn unlock lower sample complexity and help close the gap to batch-RL%
\footnote{Following terminology in \cite{elsayed2024streamin}, not to be confused with offline-RL.}
algorithms.
\cite{elsayed2024streamin} and \cite{elelimy2025gradient} both highlight the importance of eligibility traces among the other algorithmic novelties they employ, such as sparse initialisation of weight, novel regularized objectives and optimisers, as well as liberal use of normalisation for observations, rewards and activations, in breaking the ``streaming barrier'' towards an efficient online learner.

It has been remarked however that eligibility trace methods bear more than a passing similarity to \ac{sgdm} \citep{polyak1964sgdm}, even as recent as \cite{elsayed2024streamin}.
Furthermore, optimisation algorithms that also implement running averages of the derivative of the objective such as \ac{adam}~\citep{kingma2015adam} and to a lesser extent \ac{rmsprop}~\citep{tieleman2012rmsprop}, have been critical to training \ac{rl} agents with neural-network function approximation.
In this light, we aim to revisit whether:

\begin{mdframed}[style=leadingQ]
  \hangindent=2em\hangafter=1
  Q1: Are objectives and optimisers developed for batch reinforcement learning competitive in the online setting?
\end{mdframed}

Deploying a careful and extensive empirical protocol, we find that objectives, update rules and optimisation routines commonly used in batch-RL, yield surprisingly strong results once adapted to the streaming setting.
These strong empirical findings should contribute on their own to shaping the discussion around the somewhat nascent field of streaming \textit{deep} reinforcement learning.
They hint towards the generality and robustness of \tdzero\ methods when coupled with \adam--style updates.
And should establish stronger baselines for the development of new streaming algorithms.

\begin{mdframed}[style=leadingQ]
  \hangindent=2em\hangafter=1
  Q2: What explains the performance of \tdzero\ methods and \adam--style updates?
\end{mdframed}
Key for obtaining these results is the peculiar set of hyperparameters we settled on for \adam.
Our top configurations invariably required a coefficient of the running average of the gradient close to 1.0 and a very large value for the numerical stability term.
In section \ref{sec:results} we test several hypotheses that could explain this choice of values.
Similarly, we notice a preference for algorithms that have a bounded derivative of the objective, such is the case of \ac{c51} and some \ac{dqn} versions.

Finally, we distil these observations and our analysis into an algorithm based on eligibility traces we term Adaptive Q($\l$) that improves on existing methods.
To conclude this section we summarise the contributions we bring:
\begin{itemize}
  \item \textbf{The unreasonable effectiveness of \adam.} We show evidence that objectives and optimisers developed in the batch-RL setting are very strong performers in the streaming protocol once they are well tuned.
  \item \textbf{Mechanistic insights.} We identify and discuss several important properties for streaming RL: the variance-adjusted updates based on long gradient histories, the role of bounded derivatives of the objective, and the large effects of $\varepsilon$ on performance.
  \item \textbf{Adaptive Q$(\l)$.} We propose a new eligibility-trace update rule that improves on existing methods.
  \item \textbf{Extensive benchmarking.} We present a thorough empirical study, evaluating \dqn\ and \cdqn\ which we adapt to the streaming setting, alongside \streamq, on \numGames\ Atari games.
\end{itemize}

\section{Background}
\label{sec:background}

This work is set in the classical episodic reinforcement learning setting and, for convenience follows the conventions defined by \cite{elsayed2024streamin}, \cite{elelimy2025gradient} and others: an \emph{agent} interacting with an \emph{environment} generates a time-indexed sequence $S_0, A_0, R_1, S_1, A_1, R_2, ..., S_T$, by following its state-conditioned action generating behaviour, the policy $A_t\sim\pi\bp{\cdot\vert S_t}$.
Its objective is to maximise the sum of future discounted return $G_t \doteq \sum^T_{k=t+1} \g^{k-t-1} R_k$, with $\g$ controlling how much weight to put on the most immediate rewards as opposed to the ones received later in the trajectory.
It does so by estimating the expected return of being in state $S_t=s$ and following the policy $\pi$, which we call the value function $v_{\pi}(s) \doteq \E[\pi]{G_t | S_t = s}$.
When the estimator of $v_{\pi}$ uses function approximation, we denote it $v(s,\wt)$, with $\wt$ a parameter vector.
Generally, we don't only care about estimating the utility of a state, also known as the \emph{value prediction} problem.
Moreover, we are interested in its counterpart, the \emph{control} problem, with the corresponding action-value function $q_{\pi}(s,a) \doteq \E[\pi]{G_t \vert S_t=s, A_t=a}$ and its function approximation equivalent $q(s,a,\wt)$.
Optimising the objective in the control problem is then a matter of finding the optimal policy $\pi^{\star}$ that maximises the state-action value function: $q_{\pi}(s,a) = \max_{\pi} q_{\pi}(s,a)$.
\vspace{-\baselineskip}
\paragraph{Temporal difference learning.}
The straightforward approach to estimating the expected discounted sum of rewards is to wait till the end of the episode and compute it for every preceding state.
This estimate of the return makes for a learning target towards which we can adjust the current value function: $\wt_{t+1} \doteq \eta \left(G_t - v\left(S_t,\wt_t \right)\right)\nabla_{\wt} v(S_t,\wt_t)$.
The downside of Monte-Carlo (MC) methods is the policy remains unchanged until the end of the episode.
One way towards more frequent updates is to compute a learning target based on the estimator we are training, evaluated at the very next state, and the reward we received by taking an action at the current step: $R_{t+1} + \g v(S_{t+1}, \wt)$.
The resulting objective function to minimize then becomes the temporal-difference (TD) error: $\delta \doteq R_{t+1} + \g v(S_{t+1}, \wt) - v(S_t, \wt)$, of which we will make heavy use in this work.
For completion, the TD error relevant for the control problem is $\delta \doteq R_{t+1} + \g \max_a q(S_{t+1}, a, \wt) - q(S_t, A_t, \wt)$.
Because the target uses the prediction at the immediate next state $S_{t+1}$, we call it a TD(0) method.
But we can bootstrap the learning target farther into the future, giving way to $n$-step methods, where the return is estimated by:
% careful with the use of comment below
$G_{t:t+n} \doteq%
%R_{t+1} + \g R_{t+2}+...,+\g^{n-1} R_{t+n}+\g^n v(S_{t+n},\wt)%
\sum_{k=0}^{n-1} \g^k R_{t+k+1} + \g^n v(S_{t+n},\wt)
$.
Choosing $n$ allows for interpolating between the two main estimators so far: the one-step look-ahead
%
% \Florin{should check whether ``look-ahead'' is reserved for dynamic programming/search}
%
in TD(0) and MC.

\vspace{-\baselineskip}
\paragraph{On $\lambda$-returns.}
The vast majority of batch-RL algorithms are squarely set in the \emph{forward}-view we just described.
For a given state the agent ``looks'' forward in time at the future rewards received by the current policy and decides how to update the value estimate based on them.
This works out nicely for TD(0) and small values of $n$ for $n$-step return methods, but quickly becomes tricky to implement efficiently for other multistep methods, especially in conjunction with resampling strategies \citep{daley2019reconciling}.
One such alternative is the $\l$-return, an estimator of the return that further balances the bias-variance tradeoff by computing an weighted average over all $n$-step returns along a trajectory, $G^{\l}_t \doteq (1-\l) \sum_{n=1}^{T-t-1} \l^{n-1} G_{t:t+n} + \l^{T-t-1} G_t$.
Setting $\l=0$ recovers TD(0), while with $\l=1$ the estimator becomes the MC return.
This estimator is rarely encountered in pure value-based methods with neural network approximation because it requires recomputing each $n$-step return every time.

\vspace{-\baselineskip}
\paragraph{Eligibility traces.}
$\lambda$-returns can however be efficiently implemented if we take on the $backward$-view where each update to the value function depends on the current TD-error and some running statistic of past events.
The key insight is to have a \emph{trace} vector $\zt_t$ that mirrors and is the size of $\wt$.
While $\wt$ stores the long-term coefficients required for estimating $v_{\pi}(s)$, the role of $\zt$ is to record whenever a component of $\wt$ was sensitive in producing an estimate.
This record of the parameter ``activity'' is decayed towards 0 by $\lambda$ and reset in terminal states.
If the TD-error is non-zero, then $\wt$ will be updated according to the values of the error and the corresponding components of the eligibility vector $\zt$.
Specifically, if we denote the sensitivity%
% \Florin{is this the right word?}
%
of the value function with respect to the weights $\gt_t \doteq \nabla_{\wt} v(S_t,\wt)$ and initialise $\zt_0 \doteq \bm{0}$, then the update becomes:
\begin{align}
  \zt_t &\doteq \g\l \zt_{t-1} + \gt_{t}\\
  \wt_{t+1} &\doteq \wt_t + \eta \delta_t \zt_t
  \label{eq:et}
\end{align}

With linear function approximation the eligibility trace update in Eq.~\eqref{eq:et} can be equivalent to those of forward-view algorithms implementing $\l$-returns \citep{sutton1988tdlmbda,seijen2014trueOnlineTDlmbda}.

\vspace{-\baselineskip}
\paragraph{Objectives.} Generally, in the forward-view, we set the TD-error objective for minimisation by formulating it as a mean squared error (MSE) function, $\Lc(\wt) \doteq \delta^2$ and optimising it with semi-gradient descent.
Very often in practice MSE is replaced by SmoothL1Loss, $L_{\kappa}(\delta) = \ind_{|\delta| < \kappa} \left( 0.5\delta^2/\kappa \right) + \ind_{|\delta| \geq \kappa} \left( |\delta| - 0.5\kappa \right)$.
For $\k=1$ it smoothly transitions from the squared loss to the mean absolute error as the magnitude of $\delta$ increases.

\vspace{-\baselineskip}
\paragraph{Variance-adjusted optimisation methods.}
Most consequential to achieving the first strong results on challenging control problems using neural networks was the use of a new class of optimisation algorithms that divides the derivative of the objective function or an estimate of its first moment, by an estimate of the second moment.
An early example was the use of \rmsprop\ \citep{tieleman2012rmsprop} for successfully training \ac{dqn} \citep{mnih2015dqn}.
To our knowledge, \cite{bellemare2017distributional} introduces the use of \adam\ in \ac{dqn}-style methods and \cite{hessel2017rainbow,ceron2021revisiting} confirm its advantages over \rmsprop\ on a variety of agents.
In this work we refer as \emph{variance-adjusted methods} to algorithms that use the second moment of the gradient to scale the update.
\adam\ in particular approximates the first and second moments using $\beta$-normalised exponential moving averages.
Ignoring bias-correction and letting $\gt_t \doteq \nabla_{\wt} \delta_t^2$, the update is:
\begin{align}
  \mt_t   &\doteq \beta_0 \mt_{t-1} + (1-\beta_0) \gt_t      & \rhot &\doteq \mt_t/(\sqrt{\vt_t} + \varepsilon) \\
  \vt_t   &\doteq \beta_1 \vt_{t-1} + (1-\beta_1) {\gt_t}^2    & \wt_t   &\doteq \wt_{t-1} - \eta \rhot
  \label{eq:adam}
\end{align}
Comparing to Eq.~\eqref{eq:et}, we notice that both compute a running statistic of the gradient with respect to $\wt$.
However, whereas $\zt$ accumulates the derivative of $v(S_t,\wt)$ and is reset periodically, $\mt$ is a long-running average of the derivative of the objective and is generally not reset.
While both are running statistics, their roles and dynamics are different.
The eligibility trace is intended to accumulate the signal and tends to increase in magnitude, dampen oscillations and speed up learning, not unlike \ac{sgdm} \citep{polyak1964sgdm,rumelhart1986backprop}, which was already pointed out by \cite{elsayed2024streamin}.
\adam's EMA instead approximates the expected value of the gradients and tends to stay at the same scale as the signal.

\vspace{-\baselineskip}
\paragraph{Distributional RL.}
Rather than estimating only the expected return, distributional reinforcement learning \citep{bellemare2017distributional} models the full distribution of the return $G_t$, represented by the random variable $Z(s, a, \wt)$ such that $q(s, a, \wt) \doteq \E{Z(s, a, \wt)}$.
%
%This approach captures the inherent aleatoric uncertainty of the environment.
%
One way to approximate it is using a discrete set of $K$ fixed atoms $\{z_i\}^K_{i=1}$ spanning a specific range of possible returns, where the model estimates the probability mass $p_i(s, a, \wt) \doteq \text{Pr}\bp{Z(s, a, \wt) = z_i}$ associated with each atom.
The learning objective then is to minimize the Kullback-Leibler divergence between the current distribution and a target distribution formed by the sample $R_{t+1} + \g Z(S_{t+1}, A^\star, \wt)$, where $A^\star \doteq \arg\max_a q(S_{t+1}, a, \wt)$, yielding the \ac{c51} algorithm.
%
% Alternatively, Quantile Regression DQN (QR-DQN) \citep{dabney2018qr}, models the distribution using $N$ atoms with fixed, uniform probabilities $1/N$, while learning their locations $\theta_i(s, a, \wt)$ instead.
%which correspond to the quantiles of the return at targets $\tau_i = \frac{2i-1}{2N}$.
%
% The practical objective in QR-DQN is then the minimisation of the quantile Huber loss.

\section{Empirical setup}   % ----------------------------------------------------------
\label{sec:setup}
\paragraph{Adapting old algorithms to new setups.}
We closely follow the setup introduced by \cite{elsayed2024streamin}, using the environment and normalisation wrappers provided by the authors, with no modification, for all the algorithms studied here.
Our implementation of \acs{sq} closely follows the reference, and we were able to confirm that it performs on par with the original work in \minatar\ and the eight \atari\ environments the authors originally evaluated it.

We adapt \ac{dqn} and \ac{c51} to the online setting by removing the target network and the replay buffer.
Distributional algorithms such as \ac{c51} \citep{bellemare2017distributional} often have a much larger final layer, the size of $A \times K$, where $K$ is the number of bins or quantiles.
In order to compare estimators of the same capacity, in all our experiments we scale the layer before the output for non-distributional methods such that the number of weights is similar for all algorithms.
%
% Tbl.~\ref{tab:agent-summary} summarises the changes.
%
\begin{table}[hbp]
  \caption{Changes to \ac{dqn} and \ac{c51} for the online setting}
  \vspace{-1em}
  \begin{center}
    \begin{tabular}{ll}
      \multicolumn{1}{l}{Preserved from \acs{sq}}  &\multicolumn{1}{l}{Other changes}
      \\ \hline
      $\triangleright$ Sparse initialisation of weights
      &
      $\triangleright$ No target network\\
      $\triangleright$ Observation and reward normalisation
      &
      $\triangleright$ No experience replay\\
      $\triangleright$ Architecture and liberal use of LayerNorm
      &
      $\triangleright$ Equalized network capacity\\
    \end{tabular}
  \end{center}
  \label{tab:agent-summary}
  \vspace{-\baselineskip}
\end{table}

For clarification, we note here that \cite{elsayed2024streamin} also adapt \dqn\ to the streaming setup by setting the replay buffer to a size of $1$, along with an update frequency of $1$, and call it DQN1.
We will refer to this agent several times in this work.
In comparison, our \dqn\ implementation also removes the target network, uses the same architecture and initialisation as \streamq\ and the same environment wrappers.

\vspace{-\baselineskip}
\paragraph{Benchmarks and hyperparameter tuning sets.}
\label{par:benchmarks}
For our empirical study we settle for \minatar\ \citep{young2019minatar} and \ac{ale} \citep{bellemare2013ale}, two of the better established benchmarks in discrete control.
For the \minatar\ experiments we always train on all five games included in the benchmark.
Although \cite{ceron2021revisiting} establishes \minatar\ as a good predictor of \ac{ale} performance, our experiments suggest significant differences in algorithm rankings in the online setting.
Previous efforts in the online setting \citep{elsayed2024streamin,elelimy2025gradient}  evaluated their proposals on a limited subset of \atari\ games.
While we were not able to train on the full 57 game set either, we take further steps to make sure our results are predictive for the entire benchmark.
\cite{aitchison2023atari5} studies how predictive is the performance on each \atari\ game to the entire \ac{ale} benchmark, and it further identifies the most predictive subset of 5 games, called \textsc{Atari-5}.
We use this subset for all our hyper-parameter searches, with one change: we replace \textit{DoubleDunk}, in which all agents flat-line, with the most predictive game not in the original 5-set, \textit{Amidar}.
We call this subset \atariFive.
The \numGames\ games in the main comparison are also selected based on \cite{aitchison2023atari5}, in the order of their correlation to the aggregate performance.

\vspace{-\baselineskip}
\paragraph{Evaluation.}
We perform 7 training runs with different initialisations for each game in the main \atari\ experiment and use 9 seeds for the \minatar\ experiments.
Hyperparameter tuning was performed with 3 to 5 runs.
When tuning on \atariFive\ we only train for $12.5\text{M}$ steps making for $25\%$ of the usual budget.
Reporting the performance we follow \cite{agarwal2021rliable} in using stratified bootstrapping for confidence interval estimation, interquartile mean for aggregates, along with estimating the probability of improvement.

\section{Are established objectives and optimisers effective in streaming RL?}
\label{sec:results}

We developed our initial observations and intuitions on the \minatar\ environments.
A relatively sparse grid-search on a subset of games with \dqn\ and \cdqn\ revealed two important findings that we kept confirming for the rest of the project.
First, that both algorithms gain in performance with increasingly large values of $\beta_0$, the coefficient that controls the ``length'' of the history of gradients in \adam's exponential moving average of the gradient.
The default value, of $\beta_0=0.9$ is used throughout batch-RL, and defines a relatively small sliding window, which ensures the trace captures gradient information from the most recent several tens of steps.
We settled for $\beta_0=\beta_1=0.999$ for all our experiments instead.
Second, we noticed a strong increase in performance with larger values of $\varepsilon$.
We defer the discussion of this choice of hyperparameters to Sec.~\ref{sec:analysis}.

\begin{wrapfigure}{r}{0.5\textwidth}
  \vspace{-\baselineskip}
  \begin{center}\includegraphics{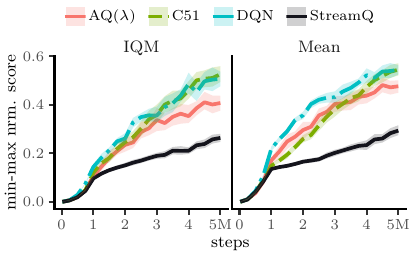}\end{center}
  \vspace{-\baselineskip}
  \caption{
    Aggregated normalised score on five \minatar\ environments.
    Classical RL objectives using \adam\ are strong performers.
  }
  \label{fig:ma-tc-iqm-mean}
  \vspace{-\baselineskip}
\end{wrapfigure}

Fig.\ref{fig:ma-tc-iqm-mean} illustrates the final results on the five \minatar\ environments.
Surprisingly, \dqn\ and \cdqn\ both outperform \streamq.
Significant is the performance of \dqn\ when compared to DQN1 from \cite{elsayed2024streamin} which always underperforms \streamq, given that they share the optimiser and the objective function.
We explain this difference because of our decision to level the field by allowing all methods to use the environment wrappers and the same estimator architecture, as well as to our specific choice of hyperparameters for \adam.

Moving on to the \numGames\ \atari\ games selected as described above, the conclusions change to some degree and Figs.~\ref{fig:main} and~\ref{fig:ale-tc-mean-grid} paint a more nuanced view.
Whereas DQN1 was failing catastrophically in the eight \atari\ games selected by the authors, ours showcases a decent performance, although not to the level of \streamq.
However, \cdqn\ manages to raise to the level of \streamq\ and is given a $50\%$ chance of improving over it by the Mann-Whitney U-statistic \citep{mann1947test,agarwal2021rliable}.

\emph{Given these results, we must answer the original question in the affirmative.
The classical objective functions developed for batch-RL, when coupled with a properly tuned \adam, demonstrate a reasonable performance in the two benchmarks we evaluated on.}

\section{What drives the performance of established objectives with \adam?}
\label{sec:analysis}

Undoubtedly, one of the main reasons we are able to train these classic agents in the streaming setup in the first place, are the normalisation techniques and the weight initialisations introduced by \cite{elsayed2024streamin}.
The separate impact of reward, observation and activation normalisation and the sparse weight initialisation scheme, have been already thoroughly ablated in prior work, and we refer the reader to it.

We turn, in what follows, on what we believe are some of the factors that make these classical algorithms perform well in the streaming setup when compared to \streamq, which, in addition, employs eligibility traces and a novel optimisation algorithm that avoids updates that could overshoot the TD target, Overshooting-bounded Gradient Descent or ObGD.

\begin{figure}[h]
    \begin{center}
      \includegraphics{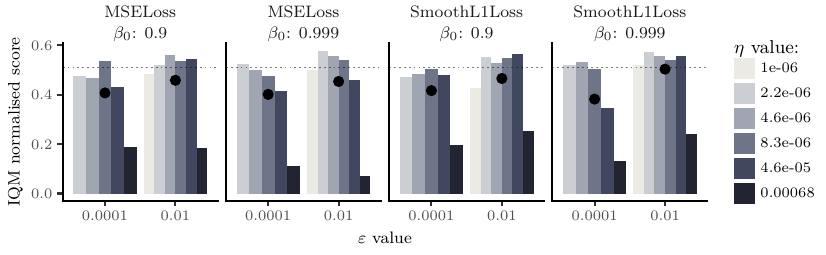}
    \end{center}
    \caption{
    On aggregate, \dqn\ performs better with larger $\beta_0$ values and bounded objectives such as SmoothL1Loss.
    $\bullet$ denotes the IQM score over $\eta$ values given a value of $\varepsilon$.
    Each bar is the IQM of 5 games $\times$ 3 seeds $\times$ last 10 evaluations.
    }
    \label{fig:ale-dqn-eps-barplot}
\end{figure}

\paragraph{Bounded objectives.}
The choice of objective function profoundly impacts optimisation stability in reinforcement learning, especially under the severe non-stationarity of the streaming setup.
MSE is affected by unbounded gradients with respect to predictions.
Consequently, when confronted with large Temporal Difference (TD) errors, the linear derivative of MSE translates these errors into massive, destabilizing gradient steps.
Recent results by \cite{palenicek2025xqc} demonstrate that bounding gradient norms is critical for maintaining a stable effective learning rate under non-stationary targets and bootstrapping.
Crucially, employing an objective function with strictly bounded gradients allows for the effective gradient update to be theoretically upper bounded.
As shown by \cite{palenicek2025xqc}, the Cross Entropy loss implemented by \cdqn\ is representative for this type of objective with bounded derivative, and we believe much of the performance it demonstrates in \atari\ (Fig.~\ref{fig:main}) and \minatar\ (Fig.~\ref{fig:ma-tc-iqm-mean}) is because of this property.
An increase in performance is also shown in Fig.~\ref{fig:ale-dqn-eps-barplot}, when changing the objective function from MSE to SmoothL1Loss.
% By clamping the derivative for large errors, a bounded objective acts as a safeguard against outliers, safely restricting the maximum step size and ensuring continuous, stable learning without the risk of divergence.
%

\begin{figure}[h]
    \begin{center}
      \includegraphics{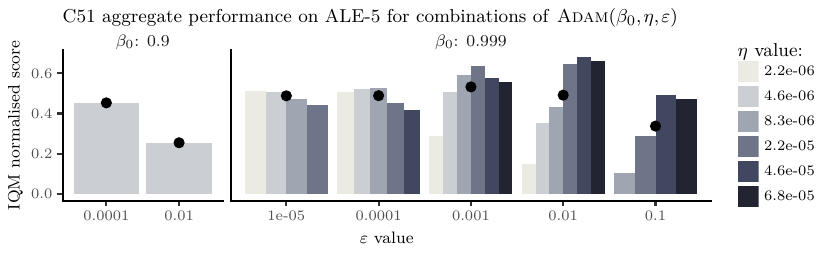}
    \end{center}
    \caption{Performance increases with higher values of $\varepsilon$ and there is no obvious scaling of the step size $\eta$ that can compensate for it.
    $\bullet$ denotes the IQM score over $\eta$ values given a value of $\varepsilon$.
    Each bar is the IQM of 5 games $\times$ 3 seeds $\times$ last 10 evaluations.
    }
    \label{fig:ale-c51-eps-barplot}
\end{figure}

\paragraph{The longer the history of the gradients, the better.}
We explain in Sec.~\ref{sec:background} that both \adam\ and eligibility traces include mechanisms for storing gradient statistics.
Although different in intent and dynamics, \adam's $\mt$, like the eligibility vector $\zt$ accumulates gradient information with a window size determined by $\beta_0$.
It maybe came as no surprise that in our initial experiments with \minatar\ we noticed a strong preference towards large $\beta_0$ values, beyond the usual $\beta_0=0.9$ found in the literature.
In Fig.~\ref{fig:ale-dqn-eps-barplot}, on the \atariFive\ subset, we notice a similar trend, where the robust median (IQM) over learning rates is higher for $\beta_0=0.999$.
Similarly, in Fig.~\ref{fig:ale-c51-eps-barplot}, on the same \atariFive, we plot on the left the two $(\varepsilon, \eta)$ combinations that performed better for $\beta_0=0.9$ and observe that they perform worse than the ones with $\beta_0=0.999$.

\paragraph{\acs{adam}'s $\varepsilon$ as a step-size scaling factor.}
The update in Eq.~\eqref{eq:adam} would suggest that the increased performance as $\varepsilon$ grows is an artefact of the interplay between the numerical stability term and step-size $\eta$.
Indeed, in the $\eta/(\sqrt{\vt} + \varepsilon)$ relation, as $\varepsilon$ increases, the effective step size gets smaller, assuming $\sqrt{\vt}$ constant.
Note also that the updates scale higher with decreasing $\varepsilon$ as the value of $\sqrt{\vt}$ decreases.
From this we could argue that increasing $\varepsilon$ the size of the update decreases with the possible effect of a more stable (and slow) optimisation process.
However, a second intuition is that this scaling effect should be compensated by picking a different step size $\eta$.

In order to test this hypothesis, we set the following experiment.
We train \ac{c51} agents on the \atariFive\ subset for $12.5$M steps, amounting to $25\%$ of the standard training run.
For five values of $\varepsilon \in \{\num{1e-5},...,\num{1e-1}\}$ we do a grid search over $\eta$ values.
The bounded derivative of the \ac{c51} objective allows us to ignore possible stabilisation effects induced by $\varepsilon$ in the presence of outlier target values and further isolate the relative interplay between $\eta$ and $\varepsilon$ if it exists.
Indeed, for this hypothesis to hold fully, we would expect a step-size $\eta$ that performs equally well for each $\varepsilon$ value.
We observe in Fig.~\ref{fig:ale-c51-eps-barplot} that indeed there is an optimal step size for each $\varepsilon$ value.
However, the performance increases with $\varepsilon$ instead of being constant at this optimal step size, therefore invalidating our initial hypothesis.
An even clearer picture of the requirement to carefully tune $\varepsilon$ is observed in Fig.~\ref{fig:ma-qr-eps-barplot}, the result of a grid-search on \minatar\ using \ac{qr}, another distributional algorithm, that implements Huber quantile loss, also an objective with a bounded derivative. Details are available in App.~\ref{app:minatar}.
We must conclude that the performance excess cannot be fully explained away by a proportional scaling of the step size and an additional mechanism must be at play.

\paragraph{\ac{adam}'s $\varepsilon$ is an SNR filter.}%
\ac{rl} has the particularity of having to deal with a long tail of features that are rarely encountered by the agent: entities that appear only sparsely or after bottleneck states, objects and tokens that are relevant only to certain parts of the game.
It is justified to assume these features are correlated with sparsely or noisily updated components of the weight vector.
Updating both the sparse and the dense or the noisy and the stable directions of the parameter space with the same scale is likely to produce estimation errors, cause instabilities, and result in the policy diverging.
Ideally, an optimiser should therefore be able to adjust the step-size of each component such that rare or noisy gradients result in conservative updates.

% % \begin{wrapfigure}{r}{0.5\textwidth}
%   \begin{figure}[h]
%   % \vspace{-\baselineskip}
%   \begin{center}\includegraphics{adam_sparse_discrete_two_eps.pdf}\end{center}
%   % \begin{center}\includegraphics{adam_sparse_continuous.pdf}\end{center}
%   \caption{%
%   Higher values of \ac{adam}'s $\varepsilon$ converge in a sparse optimisation problem.
%   %
%   Because the $w_{\text{sparse}}$ component of $\sqrt{\vt}$ is small, $\varepsilon$ dominates the $\eta\mt/(\sqrt{\vt} + \varepsilon)$ term when its value is high.
%   %
%   Therefore, it acts as a filter that scales down the gradient, producing a safe update.
%   %
%   \florin{Wrap figure later.}
%   }
%   % \vspace{-8pt}
%   \label{fig:adam-sparse-fn}
%   \end{figure}
% % \end{wrapfigure}
\begin{figure}[htbp]
  \centering
  \begin{subfigure}[b]{0.49\textwidth}
    \centering\includegraphics[width=\textwidth]{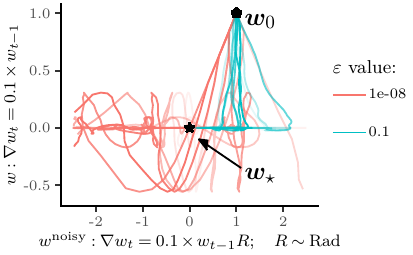}
  \end{subfigure}
  \hfill
  \begin{subfigure}[b]{0.49\textwidth}
    \centering\includegraphics[width=\textwidth]{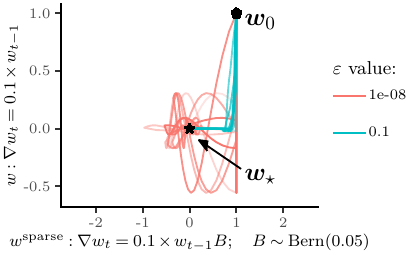}
  \end{subfigure}
  \caption{Large values for \adam's $\varepsilon$ allow for stable learning on problems which are noisy or sparse.
  The y-axis component receives a stable gradient signal in both cases, whereas on the x-axis the gradient can either change sign randomly (left) or be zero $95\%$ of the time (right).}
  \label{fig:toy}
\end{figure}

We hypothesise \ac{adam} manifests a seldom-discussed implicit mechanism for scaling the updates differently across gradient components with variable noise or sparsity characteristics.
Consider the denominator of the update, $\sqrt{\vt} + \varepsilon$ with $vt \approx \gt^2$ and ignore the bias correction.
For frequently updated or stable features, the elements $\sqrt{\vt_i}, i\in\Dc_{\text{dense/stable}}$ dominate the denominator and \ac{adam} scales down the effective step size for the directions associated with these features.
In contrast, for components of $\vt$ associated with sparse or noisy features, will result in $\varepsilon$ dominating the scaling term.
Larger values of $\varepsilon$ will be effective in this regime for bounding the maximum step size.

We propose two small toy examples to illustrate this behaviour in Fig.~\ref{fig:toy}.
Both optimisation problems depicted have just two parameters.
The component on the $y$-axis always receives a constant gradient of $0.1\times w_{t-1}$ in both problems.
In the left panel, the derivative with respect to the second weight $w^\text{noisy}$ has a similar form, but changes sign randomly at every step (on the $x$-axis).
In the right panel we formulate a sparse variation of this problem where the component on the $x$-axis is set to zero $95\%$ of the time.
Therefore, in both problems the non-constant component of $\vt$ will tend to be small, either because of noise, or because of sparse updates.

Indeed, for small values of $\varepsilon$, \ac{adam} is overly sensitive to the sparse or noisy component, amplifying the updates whenever a component of $\vt$ is small, along the corresponding dimension, resulting in unstable optimisation.
In contrast, large values of $\varepsilon$ that dominate the non-constant components of $\vt$ allows for a filtering effect on the sparse or noisy dimension, resulting in stable and early convergence.

\section{Adaptive Q$(\lambda)$}  % -----------------------------------------------------
\label{sec:aql}

Previous sections highlight the importance of i) having a long memory of the sensitivity of the value function with respect to the decisions it produced as in eligibility traces or \adam's EMA, ii) having an $\varepsilon$-modulated mechanism that adjusts the update based on the variance of this sensitivity and iii) that error signals should be bounded, like is the case of \cdqn and SmoothL1Loss.

We combine these three ideas into Adaptive Q$(\lambda)$.
Taking $Q(\lambda)$ \citep{sutton1988tdlmbda} as starting point, the algorithm uses a momentum-based eligibility trace $\zt_t$, alongside an exponential moving average of the squared gradients of the state-action value function, $\vt_t$.
We use this estimate of the second order moment of the gradient of the state-action value function to scale the eligibility trace before producing an update, similar to what \adam\ is doing.
By normalizing the eligibility trace vector $\zt_t$ with this variance estimate, we ensure that the magnitude of the updates remains stable across different parameters, even in highly non-stationary streaming environments.

We noticed however that this update was having stability issues.
Many of our training runs resulted in agents crashing because of large updates, despite the extensive normalisation techniques used in our setup.
To further improve the convergence of our method, we simply clamp $\delta$ to $(-1, 1)$, resulting in a stable algorithm across hyperparameters.
Notice how this is exactly the derivative of the SmoothL1Loss.

We reset the trace whenever the agent takes exploratory actions and when the episode ends, similar to \streamq and others in the literature.
Based on empirical experimentation, we opted not to reset $\vt$, as doing so yielded slightly worse results.
Alg.~\ref{alg:comparison} compares and summarises the updates of Q$(\lambda)$, Adaptive Q$(\l)$ and \dqn\ using \adam.
\begin{algorithm}[ht]
  \captionsetup{font=small}  % small caption
  \caption{Simplified comparison of update rules}\label{alg:comparison}
  \small  % small algo
  For all updates we require:
  $\delta \gets r + \gamma \max_a q(s',a,\wt) - q(s,a,\wt)$
  \vspace{.5em}

  \begin{minipage}[t]{0.32\textwidth}
    \begin{algorithmic}[1]
      \LineComment{\textit{$Q(\lambda)$ update}}
      \State $\gt \gets \textcolor{Cerulean}{\nabla_{\wt} q(s,a,\wt)}$
      \State $\zt \gets \gamma\lambda \zt + \gt$
      \State $\wt \gets \wt + \textcolor{OrangeRed}{\eta \delta \zt}$
      \Statex
      \Statex
      \Statex
      \If{ep. done \textbf{or} non-greedy}
      \State $\zt \gets \bm{0}$
      \EndIf
    \end{algorithmic}
  \end{minipage}
  \hfill
  \begin{minipage}[t]{0.32\textwidth}
    \begin{algorithmic}[1]
      \LineComment{\textit{$AQ(\lambda)$ update}}
      \State $\gt \gets \textcolor{Cerulean}{\nabla_{\wt} q(s,a,\wt)}$
      \State $\zt \gets \gamma\lambda \zt + \gt$
      \State $\vt \gets \gamma\lambda \vt + (1-\gamma\lambda) \gt^2$
      \State $\rhot \gets \zt/(\sqrt{\vt} + \varepsilon)$
      \State $\hat{\delta} \gets \text{clip}\bp{\delta, -1, 1}$
      \State $\wt \gets \wt + \textcolor{OrangeRed}{\eta \hat{\delta} \rhot}$
      \Statex
      \If{ep. done \textbf{or} non-greedy}
      \State $\zt \gets \bm{0}$
      \EndIf
    \end{algorithmic}
  \end{minipage}
  \hfill
  \begin{minipage}[t]{0.32\textwidth}
    \begin{algorithmic}[1]
      \LineComment{\textit{$Q$-learning update (Adam)}}
      \State $\gt \gets \textcolor{Cerulean}{\nabla_{\wt}\delta^2}$
      \State $\mt \gets \beta_0 \mt + (1-\beta_0) \gt$
      \State $\vt \gets \beta_1 \vt + (1-\beta_1) \gt^2$
      \State $\rhot \gets \mt/(\sqrt{\vt} + \varepsilon)$
      \State $\wt \gets \wt - \textcolor{OrangeRed}{\eta \rhot}$
    \end{algorithmic}
  \end{minipage}
\end{algorithm}

\aql\ delivers strong results.
In \minatar\ (Fig.~\ref{fig:ma-tc-iqm-mean}) it largely surpasses \streamq, both by the average normalised score and the robust median (IQM).
The IQM on \atari\ (Fig.~\ref{fig:main}) is largely comparable between the top three algorithms, with \aql\ having a small edge on \streamq.
The average normalised score of \aql\ however approaches three times the human baseline, in comparison to just two times the human baseline for \cdqn and \streamq.
Furthermore, the Mann-Whitney test indicates over $0.65$ probability of improvement over \streamq.

In conclusion, we believe these results validate our initial hypotheses on the importance of variance adjusted updates and bounded objectives.

\subsection{Related work}

\aql\ draws heavily from the works of \cite{javed2024swiftTD,elsayed2024streamin,elelimy2025gradient}, that sparked a renewed interest in the pure online RL setting.
It shares many of the components with \streamq.
Being a $\lambda$-return method implemented with eligibility traces, it traces its origin in the works of \cite{sutton1988tdlmbda} and \cite{mahmood2015wisTrace}.

In reasoning about the role of constraining the size of the updates, we were further inspired by \cite{javed2024swiftTD} and, specifically on the role of bounded objectives in batch-RL, by \cite{palenicek2025xqc} and \cite{farebrother2024stop}.

Among the recent advances in second-order optimisation for large language model training there's also the observation that adopting the variance-adjusting mechanism of \adam\ can still yield improvements \citep{frans2025whatMatters}, which also influenced our design decisions.

\section{Conclusion}
\label{sec:conclusion}

% \paragraph{Limitations.}
% %
% While our results establish strong baselines for streaming RL, some limitations remain.
% %
% Firstly, our current evaluation on the \ac{ale} benchmark is incomplete, encompassing just over $60\%$ of the available games.
% %
% Furthermore, the momentum parameters, $\beta_0$ and $\beta_1$, have not yet been exhaustively tuned across the \ac{ale} suite.
% %
% Finally, \aql, despite the strong performance, remains sensible to the value of $\varepsilon$, similar to \adam-based methods.

% \vspace{-\baselineskip}
% \paragraph{Future Work.}
% %
% While $AQ(\lambda)$ effectively handles expected $\l$-returns, combining the credit assignment capabilities of eligibility traces with distributional RL could provide a more comprehensive representation of the underlying return distributions, further stabilizing learning in the face of severe non-stationarity.
% %
% Furthermore, it could be promising to investigate how recent breakthroughs in deep learning optimization can be leveraged within the streaming RL setup to further increase agent performance.
%

% \vspace{-\baselineskip}
% \paragraph{Conclusion.}
%
In this work, we revisit the streaming reinforcement learning protocol to investigate whether established batch-RL techniques could be competitive.
Through extensive benchmarking across \minatar\ and \numGames\ \atari\ games, we answered this in the affirmative: when appropriately adapted, classic algorithms like \dqn\ and \cdqn\ prove to be surprisingly effective.

We then isolate some of the properties that enable this performance.
We found that the severe non-stationarity of the streaming setup requires objectives with bounded derivatives to prevent large TD errors from destabilizing learning.
Furthermore, we highlighted the necessity of variance-adjusted updates featuring a long gradient history.
Finally, we took steps toward understanding the role of \adam's $\varepsilon$ parameter, demonstrating that rather than acting as a simple step-size scalar, it functions as a Signal-to-Noise Ratio (SNR) filter that facilitates smooth convergence when processing sparse or noisy features.

Building upon these insights, we introduced Adaptive $Q(\lambda)$, a small intervention on Q($\lambda$) that synergizes variance adaptation with bounded updates to set a new high-performance baseline in the streaming regime.

% \clearpage
% Appendix ----------------------------------------------------------------------------

\appendix

\section{Appendix}
\label{sec:appendix1}

\subsection{Hyperparameters}
Table \ref{tab:hyperparameters} outlines the complete set of hyperparameter configurations used for our empirical evaluation across the ALE suite.
We detail both the globally shared settings, such as the exploration schedule and discount factor, as well as the specific optimizer and algorithmic parameters tuned for each individual method.

\begin{table}[htbp]
\centering
\caption{Hyperparameter configurations for the evaluated algorithms on the ALE suite.}
\label{tab:hyperparameters}
\begin{tabular}{lcccc}
\hline
\multicolumn{5}{c}{\textbf{Shared Hyperparameters}} \\
\hline
Discount factor ($\gamma$) & \multicolumn{4}{c}{$0.99$} \\
Exploration ($\epsilon$) start $\rightarrow$ end & \multicolumn{4}{c}{$1.0 \rightarrow 0.01$} \\
Exploration decay steps & \multicolumn{4}{c}{$2.5 \times 10^6$} \\
\hline
\multicolumn{5}{c}{\textbf{Algorithm-Specific Hyperparameters}} \\
\hline
\textbf{Parameter} & \textbf{DQN} & \textbf{C51} & \textbf{StreamQ} & \textbf{AQ($\lambda$)} \\
\hline
Optimizer & Adam & Adam & ObGD & Alg.~\ref{alg:comparison} \\
Learning Rate ($\eta$) & $2.2 \times 10^{-6}$ & $4.6 \times 10^{-5}$ & $1.0$ & $4.6 \times 10^{-4}$ \\
Optimizer $\epsilon$ & $0.01$ & $0.01$ & -- & $0.1$ \\
Adam $(\beta_1, \beta_2)$ & $(0.999, 0.999)$ & $(0.999, 0.999)$ & -- & -- \\
Eligibility Trace ($\lambda$) & -- & -- & $0.8$ & $0.8$ \\
ObGD $\kappa$ & -- & -- & $2.0$ & -- \\
Number of Atoms ($K$) & -- & $200$ & -- & -- \\
\hline
\end{tabular}
\end{table}

\subsection{Additional \minatar\ results}
\label{app:minatar}

\paragraph{Quantile Regression DQN.}
We briefly experimented in \minatar\ with an alternate distributional algorithm.
Quantile Regression DQN (QR-DQN) \citep{dabney2018qr}, models the distribution using $N$ atoms with fixed, uniform probabilities $1/N$, while learning their locations $\theta_i(s, a, \wt)$ instead.
%which correspond to the quantiles of the return at targets $\tau_i = \frac{2i-1}{2N}$.
%
The practical objective in QR-DQN is then the minimisation of the quantile Huber loss.
Although our initial \minatar\ with Quantile Regression were very promising, C51 demonstrated superior performance upon transitioning to the \atariFive\ subset.
We do however present some of these results here.

\begin{figure}[h]
    \begin{center}\includegraphics[width=\textwidth]{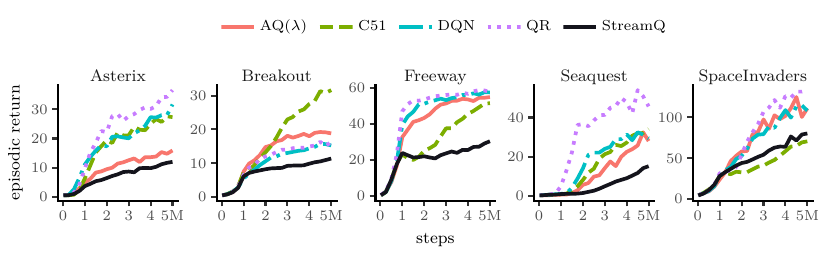}\end{center}
    \caption{\minatar\ results separated by game. The solid lines represent the mean evaluation return averaged over 9 independent runs}
    \label{fig:ma-grid}
\end{figure}
\begin{figure}[h]
    \begin{center}
      \includegraphics[width=\textwidth]{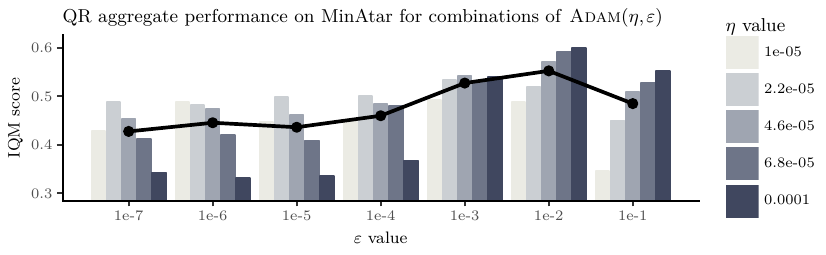}
    \end{center}
    \caption{Performance increases with higher values of $\varepsilon$ and there is no obvious scaling of the step size $\eta$ that can compensate for it. $\bullet$ denotes the mean score given a value of $\varepsilon$.}
    \label{fig:ma-qr-eps-barplot}
\end{figure}

\clearpage
\subsection{Full \atari\ results}

\begin{figure}[ht]
    \begin{center}\includegraphics[width=.97\textwidth]{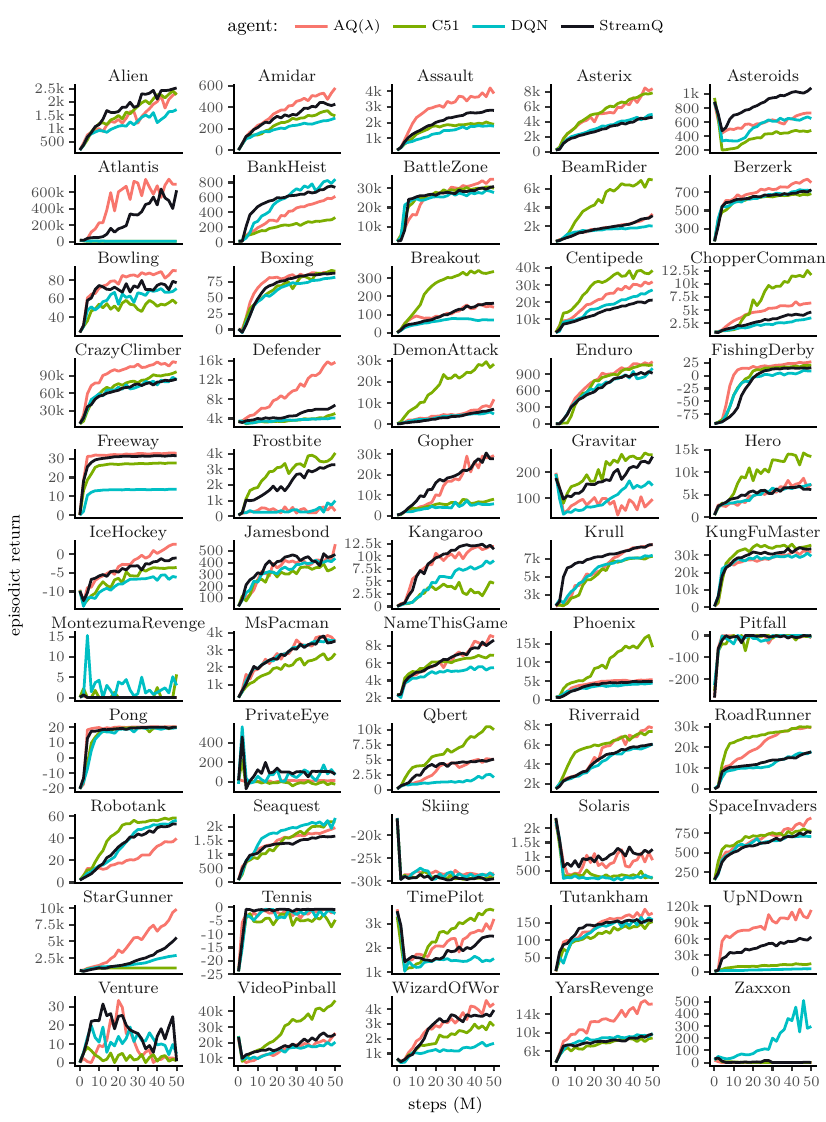}\end{center}
    \caption{\atari\ results separated by game. The solid lines represent the mean evaluation return averaged over 7 independent runs}
    \label{fig:ale-tc-mean-grid}
\end{figure}

% \section*{Appendix}
% % No label, since this can't be referenced meaningfully with \ref{}.
% This format should only be used if there is a single appendix (unlike in this document).

% \label{sec:ack}
% Use unnumbered third level headings for the acknowledgments. All acknowledgments, including those to funding agencies, go at the end of the paper. Only add this information once your submission is accepted and deanonymized. The acknowledgments do not count towards the 8--12 page limit.

\clearpage

% Bibliography ------------------------------------------------------------------------
\newpage
\bibliography{orlo}
\bibliographystyle{rlj}

% \beginSupplementaryMaterials
% % Content that appears after the references are not part of the ``main text,'' have no page limits, are not necessarily reviewed, and should not contain any claims or material central to the paper.

% \begin{figure}[ht]
%     \begin{center}\includegraphics[width=\textwidth]{ma_aql_variations.pdf}\end{center}
%     \caption{}
%     \label{fig:ma-aql-variations-barplot}
% \end{figure}

\end{document}